\documentclass[10pt,twocolumn,letterpaper]{article}

\usepackage{iccv}
\usepackage{times}
\usepackage{epsfig}
\usepackage{graphicx}
\usepackage{amsmath}
\usepackage{amssymb}
\usepackage{comment}
\usepackage{xcolor}
\usepackage{subcaption}
\usepackage{authblk}
\usepackage[accsupp]{axessibility} 
  \pdfoutput=1

\usepackage[pagebackref=true,breaklinks=true,letterpaper=true,colorlinks,bookmarks=false]{hyperref}

\iccvfinalcopy 

\def\iccvPaperID{2008} 

\ificcvfinal\fi

\begin{document}

\title{Variational Feature Disentangling for Fine-Grained Few-Shot Classification}

\author[1]{Jingyi Xu}
\author[2]{Hieu Le\thanks{Work done prior to Amazon}}
\author[1]{Mingzhen Huang}
\author[1]{ShahRukh Athar}
\author[1]{Dimitris Samaras}
\affil[1]{Department of Computer Science, Stony Brook University, NY, USA}
\affil[2]{Amazon Robotics, MA, USA}

\maketitle
\ificcvfinal\thispagestyle{empty}\fi

\begin{abstract}
Data augmentation is an intuitive step towards solving the problem of few-shot classification. However, ensuring both discriminability and diversity in the augmented samples is challenging. To address this, we propose a feature disentanglement framework that allows us to augment features with randomly sampled intra-class variations while preserving their class-discriminative features. Specifically, we disentangle a feature representation into two components: one represents the intra-class variance and the other encodes the class-discriminative information. We assume that the intra-class variance induced by variations in poses, backgrounds, or illumination conditions is shared across all classes and can be modelled via a common distribution. Then we sample features repeatedly from the learned intra-class variability distribution and add them to the class-discriminative features to get the augmented features. Such a data augmentation scheme ensures that the augmented features inherit crucial class-discriminative features while exhibiting large intra-class variance. Our method significantly outperforms the state-of-the-art methods on multiple challenging fine-grained few-shot image classification benchmarks. Code is available at: \url{https://github.com/cvlab-stonybrook/vfd-iccv21}

\end{abstract}

\section{Introduction}


Fine-grained visual data are hard to collect and costly to annotate \cite{cub,nab,dog}. 
Fine-grained visual datasets often become quite long-tailed 
and lead to classifiers overfitting to the abundant classes when trained in vanilla settings. Fine-grained few-shot learning (FSL) methods alleviate this problem since they learn discriminative class features, among visually similar classes, using as few as 5 or 1 training instances.     
   
\begin{figure}
 \includegraphics[width=\columnwidth,height=5cm]{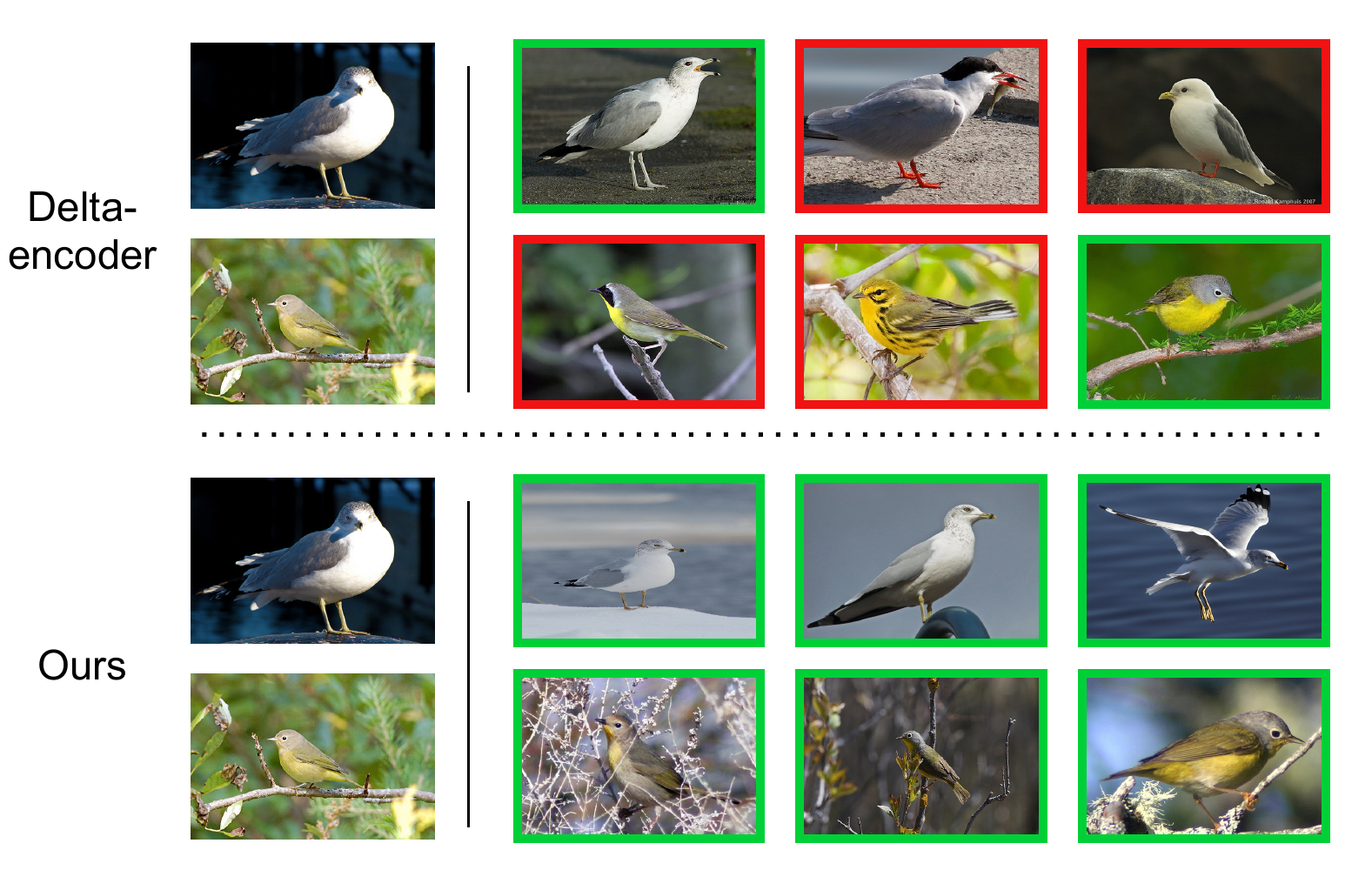}
    \centering
    \caption{\textbf{Nearest ``real sample'' neighbors of the augmented examples.} We train data augmentation methods using the base classes and search for the nearest-neighbors of the generated samples in the novel classes of the CUB dataset. The input images are shown in the first column. Each row shows the nearest neighbors of some augmented features computed from: $\Delta$-encoder \cite{delta-encoder} (1st and 2nd rows) and our method (3rd and 4th rows). Green borders indicate that the images have the same class as the input image and red borders indicate otherwise. }
 
    \label{fig:teaser}
\end{figure}

Augmenting the few-shot classes by generating additional data is a straightforward way to mitigate issues of overfitting in FSL. Nevertheless, generating diverse data reliably remains an open question~\cite{Le-etal-ICCV19,appleShrivastavaPTSW16}. 
The generated samples should contain the class-discriminative features while exhibiting high intra-class diversity.
A typical data synthesis approach is generating new samples based on adversarial frameworks \cite{metair_gan, adversarial2020kai, gao2018adversarial, metagan, dagan,Le_2020_ECCV,le2020physicsbased,m_Le-etal-ECCV18}. However, these methods suffer from a lack of diversity in the generated samples as adversarial training often mode-collapses. 
  Another approach is the feature transfer that transfers the intra-class variance from the base classes, which have many training samples, to augment features for the novel classes, in which only few samples are available \cite{delta-encoder, feature_transfer, hallucinate_features}. These methods are based on a common assumption that intra-class variations induced by poses, backgrounds, or illumination conditions are shared across categories.
The intra-class variations are either modelled as low-level statistics \cite{feature_transfer} or pairwise variations \cite{delta-encoder,hallucinate_features} and are applied directly on the novel samples. 
In this paper, we discuss two potential issues with these approaches. First, these transformations can introduce certain class-discriminative features that could alter the class-identity of the transformed features. For example, only $8.7\%$ of the augmented features using the $\Delta$-encoder\cite{delta-encoder} have their nearest ``real sample'' neighbors belong to the same classes as the original samples (see Fig. \ref{fig:teaser}). Second, 
the extracted variations might not be relevant to a specific novel sample, i.e., some bird species would never appear in sea backgrounds. Applying irrelevant variations would result in noisy or meaningless samples and degrade  classification results (see Sec. \ref{sec:generated Intra-class Variance}). These two issues are more pronounced for fine-grained classification since a small change in feature space might change the category of the feature due to the small inter-class distances. 



 
We address these issues in this paper via a novel data augmentation framework.
First, we disentangle each feature into two components: one that captures the intra-class variance, which we refer as intra-class variance features, and the other that encodes the class-discriminative features. 
Second, we model intra-class variance via a common distribution from which we can easily sample the new intra-class variations that are relevant for diversifying a specific instance. 
We show that both the feature disentanglement  
and the distribution of intra-class variability can be approximated using 
data from the base classes and it  generalizes well to the novel classes. 
The two key supervision signals that drive the training of our framework are: 1) A classification loss that ensures that the class-discriminative features contain class specific information, 2) A Variational Auto-Encoder (VAE) \cite{vae} system that explicitly models intra-class variance via an isotropic Gaussian distribution. 

Our method works especially well for fine-grained datasets where the intra-class variations are similar across classes, achieving state-of-the-art few-shot classification performances on the CUB\cite{cub}, NAB\cite{nab}, and Stanford Dogs\cite{dog} datasets, outperforming previous methods \cite{delta-encoder,meta-opt} by a large margin. We show in our analyses that the data generated by our method lies closely to the real-and-unseen features of the same class and can closely approximate the distribution of the real data. 

To sum up, our contributions are:
\begin{enumerate}
    \item We are the first to propose a VAE-based feature disentanglement method for fine-grained FSL.
    \item We show that we can train such a system using sufficient data from the base classes. We can sample from the learnt distribution to obtain relevant variations to diversify novel training instances in a reliable manner.
    \item Our method outperforms state-of-the-art FSL methods in multiple fine-grained datasets by a large margin.
\end{enumerate}



\begin{figure*}[!t]
  \centering
  \includegraphics[width=160mm]{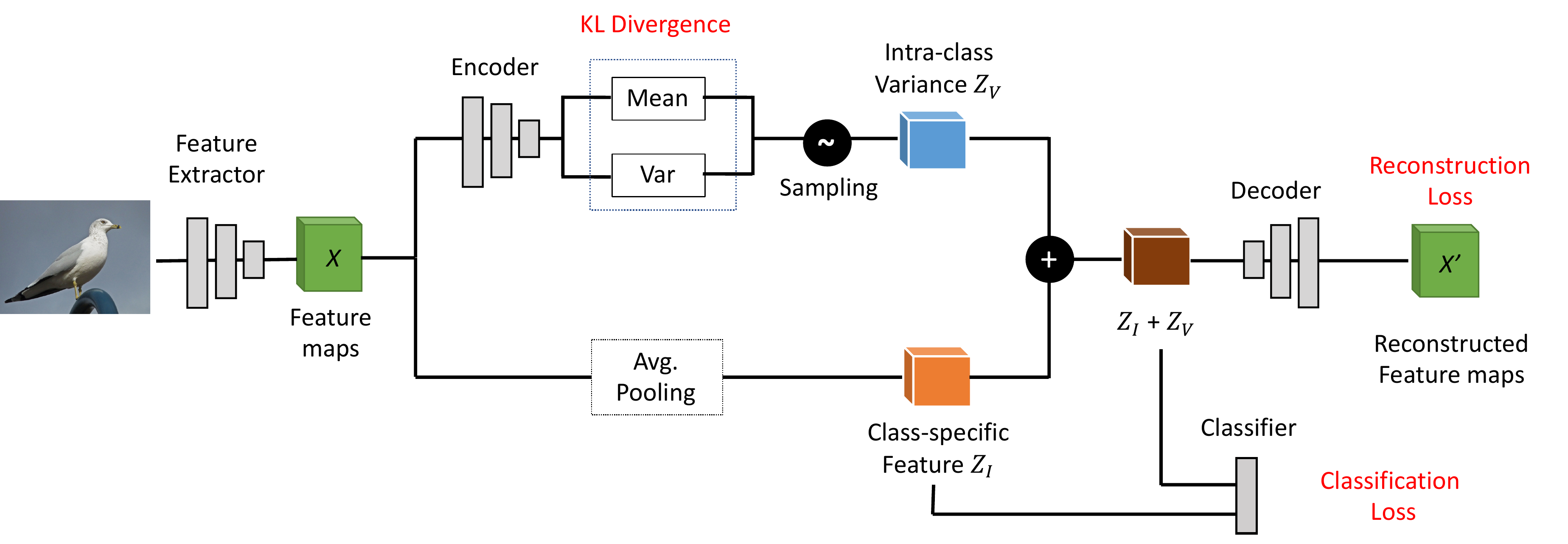}\\
   \caption{\textbf{The pipeline of our proposed method}. The input image is mapped into the image feature maps \textbf{\textit{X}}. We input \textbf{\textit{X}} into an Encoder to obtain the mean and variance of the intra-class variability distribution that are used to sample the intra-class variance feature $z_V$. The class-specific feature $z_I$ is obtained by max-pooling \textbf{\textit{X}}. $z_V$ is forced to follow an isotropic multivariate Gaussian distribution. Both $z_I$ and the combined features are used to train a classifier.
  %
   %
      %
   We sample from the learned distribution repeatedly to get multiple $z_V$ and add them to the class-specific feature $z_I$ to get the augmented features.
   These augmented features are used together with the original ones to train a more robust classifier.
  }
  \label{fig:pipeline}
\end{figure*}
\section{Related Work}

      FSL methods can be broadly organized into three categories: metric learning based, optimization based and data augmentation based.
      
      \textbf{Metric learning based methods} \cite{matchingnet,protonet,relationnet,Ye2020FewShotLV,Yang2021FreeLF,Tian2020RethinkingFI,Zhang2020DeepEMDFI} utilize the similarities between images to regularize the embedding space. 
      %
      %
      Matching Networks \cite{matchingnet} use an attention mechanism over a learned embedding of the labeled set of examples to predict classes for the unlabeled points.
      The Prototypical Network \cite{protonet} learns to classify query samples based on their Euclidean distance to prototype representations of each class.
      Sung \textit{et al.} \cite{relationnet} propose to measure the distance metric with a CNN-based relation module.

      \textbf{Optimization based methods} \cite{maml,metasgd,meta-opt,Lifchitz2019DenseCA,Santoro2016MetaLearningWM,Rajeswaran2019MetaLearningWI,Scott2018AdaptedDE} aim to design models that can generalize to new tasks efficiently.
      MAML \cite{maml} uses a meta-learner to find an initialization which can be adapted to new categories within few gradient updates using small training data. 
      Meta-SGD \cite{metasgd} learns to learn not only the learner initialization but also the learner update direction and learning rate.
      Lee \textit{et al.} propose MetaOptNet \cite{meta-opt}, which uses discriminatively trained linear predictors as base learners to learn feature representations for FSL.

      \textbf{Data augmentation based methods} \cite{dagan,delta-encoder,wang2018lowshot} generate additional training examples to alleviate the problem of data insufficiency. 
      DAGAN \cite{dagan} uses conditional generative adversarial network (GAN) to transform image features, which can be applied to novel unseen classes of data.
      Wang \textit{et al.} \cite{wang2018lowshot} propose to combine a meta-learner with a hallucinator, which can effectively hallucinate novel instances of new classes. Some methods transfer intra-class variance from the base classes to the novel class. The $\Delta$-encoder \cite{delta-encoder} extracts transferable intra-class deformations from image pairs of the same class and uses them to augment samples of the novel classes.
      Differently, our method learns a common distribution of the intra-class variability without pairing any images.
      %
     %
     
     Our method uses a VAE to model intra-class variance. VAEs have been used in FSL \cite{variational_fewshot,aligned_vae,proto_vae}.
     %
     %
     %
   Zhang \textit{et al.}  \cite{variational_fewshot} estimate a distribution for each class and compute the probability of a
     novel input to perform classification.
     Schonfeld \textit{et al.} \cite{aligned_vae} learn a shared latent space of image features and class embeddings via aligned VAEs.
     Instead of modelling the whole image feature, we only use variational inference to model the intra-class variance for our data augmentation framework. The concept of feature disentanglement using VAEs has been used in various applications such as person re-identification \cite{reid2019chanho,Zou2020JointDA}, metric learning \cite{metric_learning}, or multi-domain image translation \cite{unified2018liu,diverse2018lee}.
     We are the first to propose such a VAE-based feature disentanglement scheme for FSL problems. We show that such a model trained on base classes can be used to effectively augment data for the novel classes and significantly improve the classification results.

\section{Few-shot Learning Preliminaries}
In FSL, abundant labeled images of base classes and a small number of labeled images of novel classes are given.
  Our goal is to train a classifier that can correctly classify novel class images with the few given examples.
  The standard FSL procedure   includes a training stage and a fine-tuning stage.
  During the training stage, we use base class images to train a feature extractor and the classifier.
  Then in the fine-tuning stage, we freeze the parameters of the pre-trained feature extractor and train a new classifier head using the few labeled  examples in the novel classes . 
  In the testing stage, the learned classifier predicts labels on a set of unseen novel class images.
  
  %
Since the available samples during the fine-tuning stage are scarce and lack diversity, the learned classifier tends to overfit to the few samples and thus performs poorly on the test images.
  To address this, we augment the training samples with our proposed data augmentation method, which significantly improves the performance of the baseline.
   
  \section{Method}
  Our goal is to generate additional features of the few novel class images which contain larger intra-class variance.
  Fig. \ref{fig:pipeline} illustrates the pipeline of our proposed method.
  We decompose the feature representation of an input image into two components, the class-specifc feature $z_I$ and the intra-class variance feature $z_V$.
  $z_V$ is constrained to follow a prior distribution.
  Then we repeatedly sample new intra-class variance features $\tilde{z}_V$  from the distribution and add them to the class-specific feature $z_I$ to get augmented features.
  The augmented features are used together with the original features to train the final classifier. 
  In the following sections, we will describe how we model the distribution of intra-class variability via variational inference and how we use it to diversify samples from the novel set.
  %
  %
  \subsection{Variational Inference for Intra-class Variance} \label{variational_infer_intra_variance}
      
      Given an input image (\textit{i}), we first use a feature extractor to map it into a feature map $X^{(i)}$.
      We then compute the intra-class variance feature  $z_V^{(i)}$ and the class-specific feature $z_I^{(i)}$ from $X^{(i)}$ such that the embedding of the input image, $z^{(i)}$, can be expressed as:
      \begin{equation}
          z^{(i)} = z_I^{(i)} + z_V^{(i)}.
      \end{equation}
      %
      %
      Here we assume that the intra-class variance feature is generated from some conditional distribution $p(z_V)$ and the feature map $X^{(i)}$ is generated from some conditional distribution $p(X|z)$.
      
      The class-specific feature $z_I^{(i)}$ can be learned by minimizing the cross-entropy loss given the class label $y^{(i)}$:
      \begin{equation}
        L_{cls}(X^{(i)}) = L_{cross-entropy}\left(W(z_I^{(i)}), y^{(i)}\right)
      \end{equation}
      where $W$ is a classifier with a single fully connected layer. 

      We use variational inference to model the posterior distribution of the variable $z_V$. 
      Specifically, we approximate the true posterior distribution $p(z_{V}|X)$ with another distribution $q(z_{V}|X)$. The Kullback-Leibler divergence between the true distribution and the approximation is:

      \begin{equation} \label{KL Divergence}
          KL[q(z_{V}|X)||p(z_{V}|X)]\ = \int_Z q(Z|X) \textnormal{log} 
          \frac{q(Z|X)}{p(Z|X)}.
      \end{equation}
      
      Since the Kullback-Leibler divergence is always greater than or equal to zero, maximizing the marginal likelihood $p(X^{(i)})$ is equivalent to maximizing the evidence lower bound (ELBO) defined as follows:
      \small
      \begin{equation} \label{ELBO}
      \begin{split}
          ELBO^{(i)} = E_{q(z_{V}^{(i)}|X^{(i)})}&[\textnormal{log }p(X^{(i)}|z_V^{(i)})] \\
          &- KL\left(q(z_V^{(i)}|X^{(i)})||p(z_V)\right).
      \end{split}
      \end{equation}
      \normalsize
      %
      %
      Prior work \cite{metric_learning,feature_transfer,variational_fewshot} has shown that the distribution of intra-class variability can be modelled with a Gaussian distribution.
      Here we set the prior distribution of $z_V$ to be a centered isotropic multivariate Gaussian: $p(z_V) = \mathcal{N}(0, I)$.
      For the posterior distribution, we set it to be a multivariate Gaussian with diagonal covariance:
      \begin{equation}
        q(z_V^{(i)}|X^{(i)}) = \mathcal{N}(\mu^{(i)}, \sigma^{(i)}),
      \end{equation}
      \noindent where $\mu^{(i)}$ and $\sigma^{(i)}$ are computed by a probablistic encoder.
      With the reparameterization trick, we obtain $z_V^{(i)}$ as follows:
      \begin{equation}\label{reparameterization}
        z_{V}^{(i)} = \mu^{(i)} + \sigma^{(i)} * \epsilon, \epsilon \sim \mathcal{N}(0, I).
      \end{equation} 

      Since $z_I^{(i)}$ is deterministic given $X^{(i)}$, we have $p(X^{(i)}|z_V^{(i)}) = p(X^{(i)}|z_V^{(i)}, z_I^{(i)}) = p(X^{(i)}| z^{(i)})$.
      To estimate the maximum likelihood $p(X^{(i)}| z^{(i)})$, we use a decoder to reconstruct
      the original feature map from $z^{(i)}$ and minimize the $L2$ distance between
      the original feature map and the reconstructed one.
      
      From Eq. \ref{ELBO}, we now derive the loss function for the modeling of intra-class variance:
      \small
      \begin{equation} \label{intra-class-2}
      \begin{aligned}
        L_{intra}(X^{(i)}) &= \| X^{(i)} - \hat{X}^{(i)} \|^2 + KL\left(q(z_V^{(i)}|X^{(i)})||p(z_V)\right),
      \end{aligned}
      \end{equation}
      \normalsize
      where $\hat{X}^{(i)}$ is the reconstructed feature map synthesized from the sum of class-specific feature $z_I^{(i)}$
      and intra-class variance feature $z_V^{(i)}$ sampled from the distribution $\mathcal{N}(\mu^{(i)}, \sigma^{(i)})$.      
      
      The $L_{intra}$ loss includes two terms.
      The first term is the reconstruction term, which ensures that the encoder extracts meaningful information from the inputs.
      The second term is a regularization term, which forces the latent code, $z_V^{(i)}$, to follow a standard normal distribution.
      Here, instead of minimizing the Kullback-Leibler divergence directly, we decompose it into three terms as in \cite{beta-tv-vae}:
      
      \footnotesize
      \begin{equation}
      \begin{split} \label{kl-decompose}
       KL[q(&z_{V}|X)||p(z_{V})] = KL\left( q(z_{V},X)||q(z_{V})p(X) \right) + \\ &KL( q(z_{V})||\prod_{j}q(z_{V_j})) + \sum_{j}KL(q(z_{V_j})||p(z_{V_j})),
      \end{split}
      \end{equation}
      
      \normalsize
      \noindent where $z_{V_j}$ denotes the $j$-th dimension of the latent variable.
      
      The three terms in Eq. \ref{kl-decompose} are referred to as the \textit{index-code mutual information}, \textit{total correlation}, and \textit{dimension-wise} KL respectively.
       Prior work \cite{beta-tv-vae,Achille2018,Burgess2018UnderstandingDI} has shown that penalizing the index-code mutual information and total correlation terms leads to a more disentangled representation while the dimension-wise KL term ensures that the latent variables do not deviate too far form the prior. Similar to \cite{beta-tv-vae}, we penalize the total correlation with a weight \(\alpha\)
      and rewrite $L_{intra}$  as follows:
      
    \footnotesize
      \begin{equation}
      \begin{split} \label{intra-class-3}
        L_{intra}&(X^{(i)}) 
       = \| X^{(i)} - \hat{X}^{(i)} \|^2 + KL\left( q(z_V^{(i)},X^{(i)})||q(z_V^{(i)})p(X) \right) + \\ \alpha*& KL\left( q(z_V^{(i)})||\prod_{j}q(z_{V_j}^{(i)}) \right) + \sum_{j}KL\left(q(z_{V_j}^{(i)})||p(z_{V_j})\right). 
      \end{split}
    \end{equation}
     \normalsize 
     \noindent The combination of $L_{cls}$ and $L_{intra}$ drives the model to extract discriminative class-specific features
    $z_I^{(i)}$ and model the distribution of intra-class variability simultaneously.
        %
     
        %
\subsection{Objective Function}
        Given the distribution of intra-class variability, we can generate additional samples for the base classes during the training stage. For input image (\textit{i}) with extracted class-specific feature $z_I^{(i)}$ and  intra-class variability mean and variance $\mu^{(i)}$ and $\sigma^{(i)}$ respectively, we sample new intra-class variance features, $\tilde{z}_V^{(i)}$,   for this image from the distribution $\mathcal{N}(\mu^{(i)}, \sigma^{(i)})$ and add them to $z_I^{(i)}$ to obtain the augmented features $\tilde{z}^{(i)} = z_{I}^{(i)} + \tilde{z}_V^{(i)}$.
        We use these features to train our system using the following cross-entropy loss:
\begin{equation}
        L_{aug}(X^{(i)}) = L_{cross-entropy}\left(W(\tilde{z}^{(i)}), y^{(i)}\right)
\end{equation} 
        %
        %

      The overall loss function in the training stage is a weighted combination of the aforementioned terms:
      \begin{equation} \label{loss_function}
          L = L_{cls} + L_{intra} + \beta * L_{aug}
      \end{equation}
      
    \noindent where $\beta$ is the coefficient of $L_{aug}$.

\subsection{Diversifying Samples for Few-Shot Classes} 
\label{transfer_learning_intra_class_variance}
In this section, we discuss how to use our model to diversify samples for few-shot classes.
Our intra-class variance is modelled by an isotropic Gaussian distribution. Sampling from this distribution would result in an arbitrary intra-class variance feature. However, we conjecture that such an arbitrary feature may not be relevant for all instances, i.e., some birds never appear with a background of the sea. 
Note that here as all intra-class variations are mapped into a common continuous embedding space via variational inference and closely related or similar intra-class variations likely form local neighborhoods in the embedding space. Thus,
instead of sampling from the zero-mean and unit-variance distribution, we only sample from the mean and variance estimated directly from the conditional sample to obtain the likely relevant intra-class variations to this sample. 


Specifically, given an image of novel class (\textit{i})* with class label $y^{(i)*}$, we first extract the feature map $X^{(i)*}$, the class-specific feature $z_I^{(i)*}$, and the mean and variance of the intra-class variability distribution $\mu^{(i)*}$ and $\sigma^{(i)*}$ for this instance.
     We then generate additional features by adding the class-specific features $z_I^{(i)*}$ with a biased term sampled from the distribution of intra-class variability. 
     \begin{equation} 
        \tilde{z}^{(i)*}  =  z_I^{(i)*} +  \tilde{z}_V^{(i)*} , \tilde{z}_V^{(i)*} \sim N(\mu^{(i)*}, \sigma^{(i)*}),
     \end{equation}
     \noindent where $\tilde{z}^{(i)*}$ is the augmented feature and $\tilde{z}_V^{(i)*}$ is sampled from the posterior distribution $N\left(\mu^{(i)*}, \sigma^{(i)*}\right)$.
    By sampling from $N\left(\mu^{(i)*}, \sigma^{(i)*}\right)$ multiple times, we get multiple augmented features $\tilde{z}^{(i)*}$ that can be used to train the classifier. In Sec. \ref{sec:generated Intra-class Variance}, we verify the effectiveness of this sampling scheme.

   \begin{table*}[t] 
    \centering
  \resizebox{0.83\textwidth}{!}{%
    \begin{tabular}{l|cc|cc|cc}
      \hline
      Method  & \multicolumn{2}{c|}{CUB} & \multicolumn{2}{c|}{NAB} &
                \multicolumn{2}{c}{Stanford Dogs}  \\
      &  1-shot & 5-shot & 1-shot & 5-shot &1-shot & 5-shot \\
      \hline
      Baseline \cite{closer-look} & 63.90  $\pm$ 0.88 & 82.54 $\pm$  0.54  & 70.36 $\pm$ 0.89 & 87.91  $\pm$ 0.49  & 63.53  $\pm$ 0.89 & 79.95  $\pm$ 0.59 \\
      Baseline++ \cite{closer-look}      & 68.46  $\pm$ 0.85 & 81.02  $\pm$ 0.46 & 76.00  $\pm$ 0.85 & 90.99  $\pm$ 0.41 & 58.30  $\pm$ 0.35 & 73.77  $\pm$ 0.68\\
      MAML \cite{maml}            & 71.11  $\pm$ 1.00 & 82.08  $\pm$ 0.72 & 80.08  $\pm$ 0.93  & 88.87  $\pm$ 0.54 & 66.56  $\pm$ 0.66 & 79.32  $\pm$ 0.35\\ 
      MatchingNet \cite{matchingnet}    & 72.62  $\pm$ 0.90 & 84.14  $\pm$ 0.50 & 73.91  
      $\pm$ 0.72 & 88.17  $\pm$ 0.45 & 65.87  
      $\pm$ 0.81 & 80.70  $\pm$ 0.42 \\
      ProtoNet \cite{protonet}                      & 71.57  $\pm$ 0.89 & 86.37  $\pm$ 0.49 & 73.60  $\pm$ 0.83 & 89.72  $\pm$ 0.41 & 65.02  $\pm $0.92 & 83.69  $\pm$ 0.48\\
      RelationNet \cite{relationnet}  & 70.20  $\pm$ 0.84 & 84.28  $\pm$ 0.46 & 67.41  $\pm$ 0.82 &  85.47  $\pm$ 0.43 & 59.38  $\pm$ 0.79 & 79.10  $\pm$ 0.37\\
      MTL \cite{meta-transfer}  & 73.31  $\pm$ 0.92 & 82.29  $\pm$ 0.51 & 78.69  $\pm$ 0.78 &  87.74  $\pm$ 0.34 & 54.96  $\pm$ 1.03 & 68.76  $\pm$ 0.65\\
      $\Delta$-encoder \cite{delta-encoder}& 73.91  $\pm$ 0.87 & 85.60  $\pm$ 0.62 & 79.42  $\pm$ 0.77 &  92.32  $\pm$ 0.59 & 68.59  $\pm$ 0.53 & 78.60  $\pm$ 0.78\\
      MetaOptNet \cite{meta-opt}  & 75.15  $\pm$ 0.46 & 87.09  $\pm$ 0.30 & 84.56  $\pm$ 0.46 &  93.31  $\pm$ 0.22 & 65.48  $\pm$ 0.49 & 79.39  $\pm$ 0.25\\
      Ours  & \textbf{79.12}  $\pm$ 0.83 & \textbf{91.48}  $\pm$ 0.39 & \textbf{88.62}  $\pm$ 0.73 &  \textbf{95.22}  $\pm$ 0.32 & \textbf{76.24}  $\pm$ 0.87 &  \textbf{88.00}  $\pm$ 0.47\\
      \hline
    \end{tabular}}  
    
    \caption{Few-shot classification accuracy on the CUB ~\cite{cub}, NAB ~\cite{nab}, and Stanford Dogs ~\cite{dog}  dataset. All experiments are from 5-way classification with the same backbone network (ResNet12).
    The best performance is indicated in bold.
    }
    \label{tab:nab}%
    \vspace{-10pt}
  \end{table*}
 
  \begin{table*}[t] 
    
    \centering
  \resizebox{0.6\textwidth}{!}{%
    \begin{tabular}{l|cc|cc}
      \hline
      Method  & \multicolumn{2}{c|}{CUB} &
                \multicolumn{2}{c}{Stanford Dogs}  \\
      &  1-shot & 5-shot &1-shot & 5-shot \\
      \hline
      MatchingNet \cite{matchingnet}    & 45.30 $\pm$ 1.03 & 59.50  $\pm$ 1.01  & 35.80  $\pm$ 0.99 & 47.50  $\pm$ 1.03\\
      ProtoNet \cite{protonet}  & 37.36  $\pm$ 1.00 & 45.28  $\pm$ 1.03  & 37.59  $\pm $1.00 & 48.19  $\pm$ 1.03\\
      RelationNet \cite{relationnet}  & 58.99  $\pm$ 0.52 & 71.20  $\pm$ 0.40 & 43.29  $\pm$ 0.46 & 55.15  $\pm$ 0.39\\
       MAML \cite{maml}            & 58.13  $\pm$ 0.36 & 71.51  $\pm$ 0.30 & 44.84  $\pm$ 0.31 & 58.61  $\pm$ 0.30\\ 
      adaCNN \cite{adaCNN}  & 56.76  $\pm$ 0.50 & 61.05  $\pm$ 0.44 & 42.16  $\pm$ 0.43 & 54.12  $\pm$ 0.39\\
      CovaMNet \cite{CovaMNet}& 52.42  $\pm$ 0.76 & 63.76  $\pm$ 0.64 & 49.10  $\pm$ 0.76 & 63.04  $\pm$ 0.65\\
      DN4 \cite{DN4}  & 53.15  $\pm$ 0.84 & 81.90  $\pm$ 0.60  & 45.73  $\pm$ 0.76 & 61.51  $\pm$ 0.85 \\
      LRPABN \cite{LRPABN} & 63.63  $\pm$ 0.77 & 76.06 $\pm$  0.58  & 45.72  $\pm$ 0.75 & 60.94
       $\pm$ 0.66 \\
      MattML \cite{MattML}  & 66.29  $\pm$ 0.56 & 80.34  $\pm$ 0.30 & 54.84  $\pm$ 0.53 & 71.34  $\pm$ 0.38\\
      Ours  & \textbf{68.42}  $\pm$ 0.92 & \textbf{82.42}  $\pm$ 0.61  & \textbf{57.03}  $\pm$ 0.86 &  \textbf{73.00}  $\pm$ 0.66\\
      \hline
    \end{tabular}}  
    \caption{Few-shot classification accuracy on the CUB \cite{cub} and Stanford Dogs \cite{dog} dataset. All experiments are from 5-way classification with the same backbone network (Conv4).
    The best performance is indicated in bold.
    }
    \label{tab:cub_dog}%
  \end{table*}

    %

\section{Experiments}
  \subsection{Datasets} 
  We evaluate our method on three fine-grained image classification datasets: Caltech UCSD Birds (CUB) \cite{cub}, North America Birds (NAB) \cite{nab} and Stanford Dogs \cite{dog}.
     The CUB dataset contains 11,788 bird images from 200 bird species in total.
      Following the setup introduced in \cite{cub}, we sample the base classes from the 100 classes provided for training, and sample the novel set from the 50 classes provided for testing.
     The NAB dataset  contains 48,527 bird images with 555 classes, which is four times larger than CUB. Similar to \cite{metair_gan}, we adopt a 2:1:1 training, validation and test set split.
     The Stanford Dogs dataset is a subset of the Imagenet dataset designed for fine-grained image classification with 90 categories for training and validation and 30 testing categories.
     
     \subsection{Implementation Details}
      We conduct experiments with two architectures of our feature extractor: ResNet12 and Conv4 for fair comparisons with other methods using similar architectures.
      \textbf{ResNet12} \cite{resnet} contains 4 Residual blocks. Each residual block is composed of 3 \textit{conv} layers with 3 $\times$ 3 kernels.
      A 2 $\times$ 2 max-pooling layer is applied at the end of each residual block.  
      \textbf{Conv4} consists of 4 layers with 3 $\times$ 3 convolutions
      and 32 filters, followed by batch normalization (BN) ,
      a ReLU nonlinearity, and 2 $\times$ 2 max-pooling.
      
      The class-specific features are calculated by average-pooling the output of the feature extractor.
      The encoder consists of three \textit{conv} blocks followed by two fully-connected heads that output the $\mu$ and $\textnormal{log}\sigma^2$ respectively.
      The decoder consists of a fully connected layer followed by three Convolutional blocks.
      %
      

      %
      %
      
      \textbf{Training policies.} The whole network is trained from scratch in an end-to-end manner. In the training stage, we use
      the Adam optimizer \cite{adam_optimizer} on all datasets with initial learning rate 0.001
      . 
      We train our model for 100 epochs in total with
      a batch size of 16 and reduce the learning rate by 0.1 at the 40-th and 80-th epochs. 
      We empirically set $\alpha$ = 4 in Eq. \ref{intra-class-3} and $\beta$ = 1 in Eq.\ref{loss_function}. 
      
      We follow a standard few-shot evaluation scheme. 
      In the fine-tuning stage, we select 5 classes from the novel classes randomly. 
      For each class, we pick $k$ instances as the support set and 16
      instances for the query set for a $k$-shot task. The extracted features of all support set images along
      with the augmented features are used to train a linear classifier for 100 iterations with a batch size
      of 4. 
      For each feature extracted from a support image, we obtain five augmented features. The final
      results are averaged over 600 experiments.
      For data augmentation, we adopt random cropping, horizontal flipping and color jittering as in \cite{closer-look}. The final
      size of the input images is $84\times84$ .
      
       \subsection{Results} 
    Tab. \ref{tab:nab} summarizes the 5-way classification accuracy of  various methods using ResNet12 backbones.
    The results are obtained using the publicly available code of each method.
    Our proposed method outperforms the previous methods by a large margin for both 1-shot and 5-shot settings on all three datasets.
    Compared with the $\Delta$-encoder \cite{delta-encoder}, another data augmentation based method, our proposed method achieves $7.40\%$, $9.20\%$ and $7.65\%$ performance gain for the  1-shot setting and $5.88\%$, $2.90\%$ and $9.40\%$ performance gain for the 5-shot setting on the three datasets respectively.
It can be seen that our improvement in the 1-shot setting is more pronounced than in the 5-shot setting since the 1-shot setting is a more extreme case of data scarcity, in which augmenting the training data tends to be more useful. 
    
    We compare with methods using Conv4 architectures as the backbone networks in Tab. \ref{tab:cub_dog}. Here the majority of methods only report their results on the CUB and Stanford Dogs datasets. 
    %
    %
    %
    Our proposed method achieves state-of-the-art performance for both the 1-shot and 5-shot settings.
    Especially for the 1-shot setting, our method obtains $2.12\%$ performance gain for the CUB and $2.19\%$ gain for the Stanford Dogs over MattML \cite{MattML}, a newly proposed method that is aimed specifically at fine-grained few-shot visual recognition. 
    
    Our method also achieves competitive few-shot classification performances on non fine-grained datasets such as CIFAR-FS\cite{Bertinetto2019MetalearningWD} and mini-ImageNet\cite{matchingnet,meta_learn_lstm}. More details can be found in the supplementary material.
   
 \section{Additional Analyses}
 In this section, we provide additional experiments to clarify different aspects of our methods. 
  \subsection{Analysis on the generated intra-class variations}
  \label{sec:generated Intra-class Variance}

   We conduct a simple experiment to verify the effectiveness of our sampling method (Sec.\ref{transfer_learning_intra_class_variance}). Instead of sampling from the instance-conditioned mean and variance, we sample the intra-class variance feature from the zero-mean and unit-variance distribution.  
   
\begin{figure}[!htb]
  \centering
  \includegraphics[width=\linewidth]{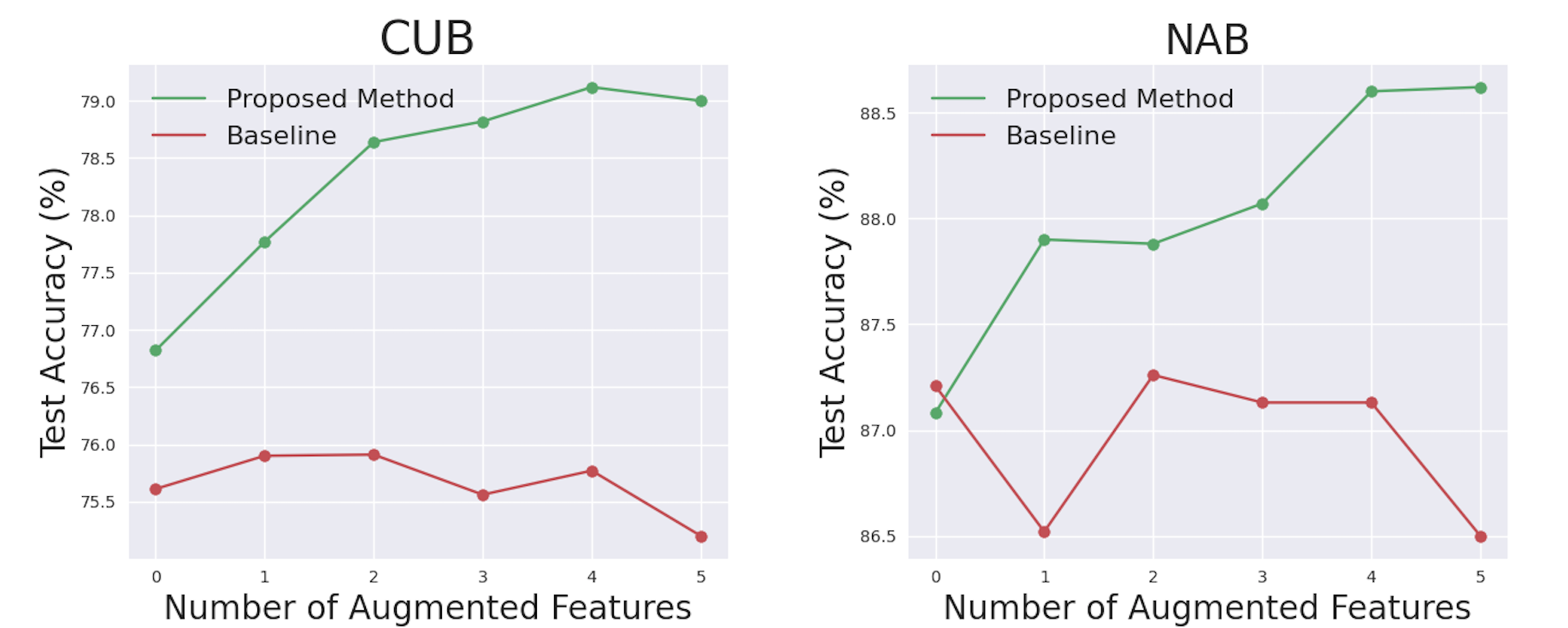}
  \caption{
  \textbf{Analysis on the generated intra-class variations.}  We augment samples with the intra-class variance features sampled from the estimated mean and variance (green lines) or from the zero-mean and unit-variance (red lines). Our sampling scheme generates features that consistently improve  classification. 
  %
  %
  %
  }\label{fig:feats_num}
\end{figure}

   Fig. \ref{fig:feats_num} summarizes the results of this experiment for 5-way 1-shot classification on the CUB and NAB dataset. As can be seen, intra-class variance features sampled from zero-mean and unit-variance do not improve the results (red lines). In contrast, our method of sampling from the instance-conditioned posterior distribution generates features that consistently improve  classification performance as the number of augmented samples increases. 

  %
  %
  %

    %
    %
    %
    %

\subsection{Comparison to other data augmentation based methods}
  
  We  compare our method with two other data augmentation based FSL methods: MetaIRNet\cite{metair_gan} and $\Delta$-encoder\cite{delta-encoder}.
  MetaIRNet uses a pre-trained image generator to synthesize additional images and combine them with the original images to form additional training samples. 
  The $\Delta$-encoder learns to synthesize transferable non-linear deformations between pairs of examples of seen classes 
  and apply these deformations to the few provided samples of novel categories.

We use the additional samples synthesized by both of these methods to train three types of classifiers:
  K-nearest neighbors (KNN), Support Vector Machine (SVM), and Logistic Regression (LR), which are then used to classify novel images. The comparisons between these methods and our method are shown in Tab. \ref{tab:comparison}.
  The superior performance of our method demonstrates that the augmented features obtained by our framework is beneficial for various types of classifiers.
  Note that for MetaIRNet \cite{metair_gan}, the results in Tab. \ref{tab:comparison} are lower than their numbers reported in the original paper since they pre-trained the backbone network on ImageNet while here all methods are trained from scratch. 
   \begin{table}[!h] 
    \centering
  \resizebox{0.48\textwidth}{!}{%
    \begin{tabular}{l|cc|cc|cc}
      \hline
      Method & \multicolumn{2}{c|}{KNN} & \multicolumn{2}{c|}{SVM} &
                \multicolumn{2}{c}{LR}  \\
      & 1-shot & 5-shot & 1-shot & 5-shot &1-shot & 5-shot \\
      \hline
      MetaIRNet \cite{metair_gan}   & 63.18  & 74.82 & 63.76 & 76.77  &  63.53 & 79.95 \\
      $\Delta$-Encoder \cite{delta-encoder}        & 67.31  & 82.67  & 76.02 & 82.87  & 76.22  & 85.17 \\
      Ours             & \textbf{75.46}  & \textbf{83.17}  & \textbf{79.07}  & \textbf{87.59}  & \textbf{78.34} & \textbf{89.30}\\ 
      \hline
    \end{tabular}} 
    
    \caption{\textbf{Analysis of different classifiers.} Few-shot classification accuracy on the CUB ~\cite{cub} dataset in 1-shot and 5-shot settings with different types of classifiers.
    }
    \label{tab:comparison}%
  \end{table}
  
  In Tab. \ref{tab:comparison_1NN}, we directly compare our method withthe  $\Delta$-encoder using K-NN classifiers (K=1). Interestingly, it can be seen that the augmented features generated using the delta-encoder decrease classification performance. In fact, we observe that the majority ($91.3\%$) of the nearest neighbors of the $\Delta$-encoder's generated features belong to different classes (some are visualized in Fig. \ref{fig:teaser}), suggesting that the pairwise transformations extracted from this method might alter the class-identities of the transformed features. On the other hand, our generated features preserve well the class identity and mildly improve the classification results.
  
   \begin{table}[!h] 
    \centering
  \resizebox{0.38\textwidth}{!}{%
    \begin{tabular}{l|cc|cc}
      \hline
      Method & \multicolumn{2}{c|}{$\Delta$-Encoder} & \multicolumn{2}{c}{Ours}   \\
      & w/o Aug & w/ Aug & w/o Aug & w/ Aug  \\
      \hline
       5-way & 69.37  & 67.31 & 74.95 & 75.46   \\
      10-way & 58.69 & 52.19  & 62.05 & 63.17   \\
      20-way & 48.10  & 38.84  & 50.19  & 50.72  \\ 
      \hline
    \end{tabular}} 
    
    \caption{\textbf{Effect of augmented features on 1NN classifier.} Few-shot classification accuracy on the CUB \cite{cub} dataset using 1NN classifier with original features vs augmented features. The original features of the $\Delta$-Encoder are from a pre-trained ResNet18 network.
    }
    \label{tab:comparison_1NN}%
  \end{table}
\def\subboxsize{50mm}
\def\W{.3\textwidth}
\def\H{40mm}
\def\Blank{5mm}
\begin{figure*}
  \centering
\includegraphics[width=\W,height=\H]{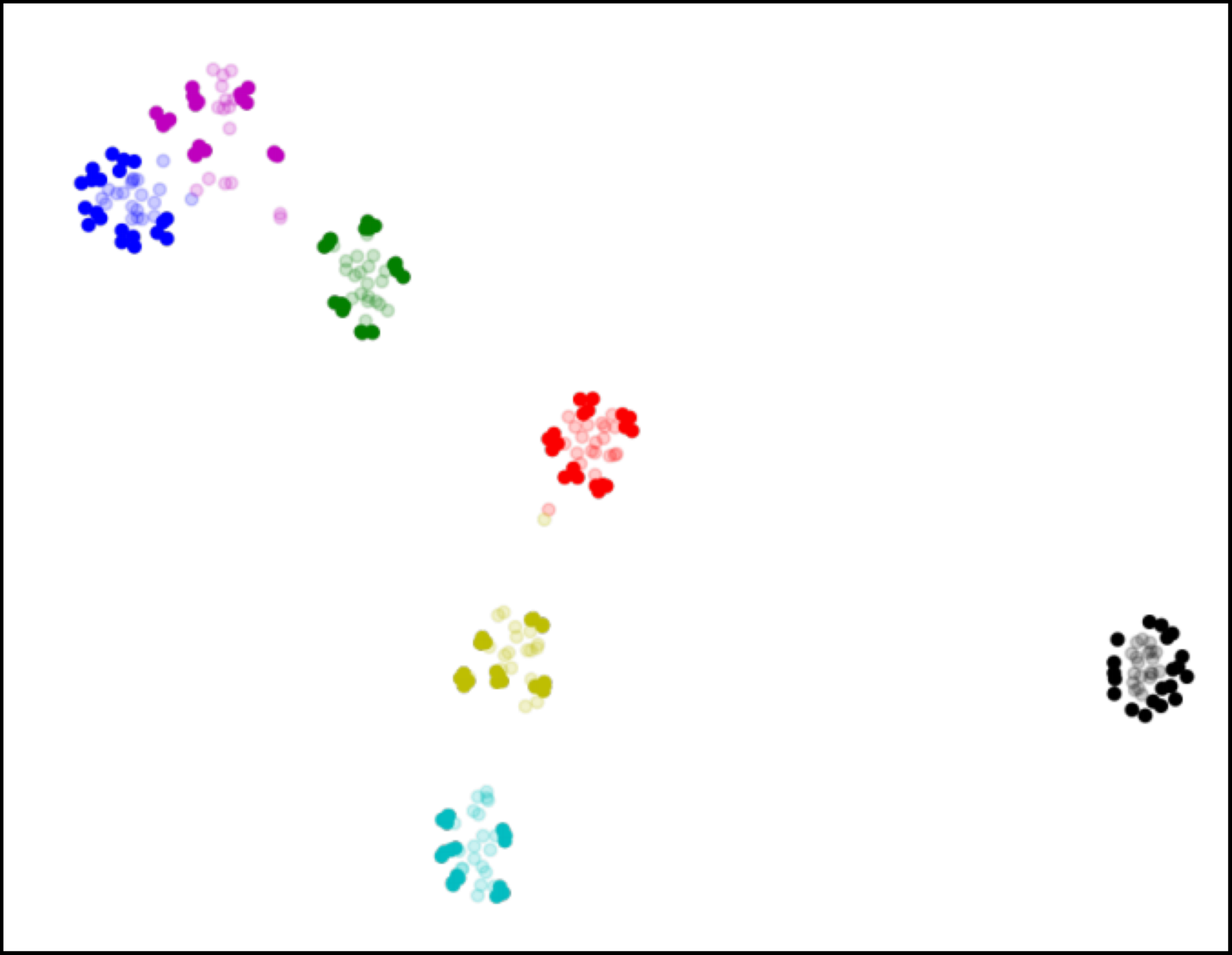}
\makebox[\Blank]{}
\includegraphics[width=\W,height=\H]{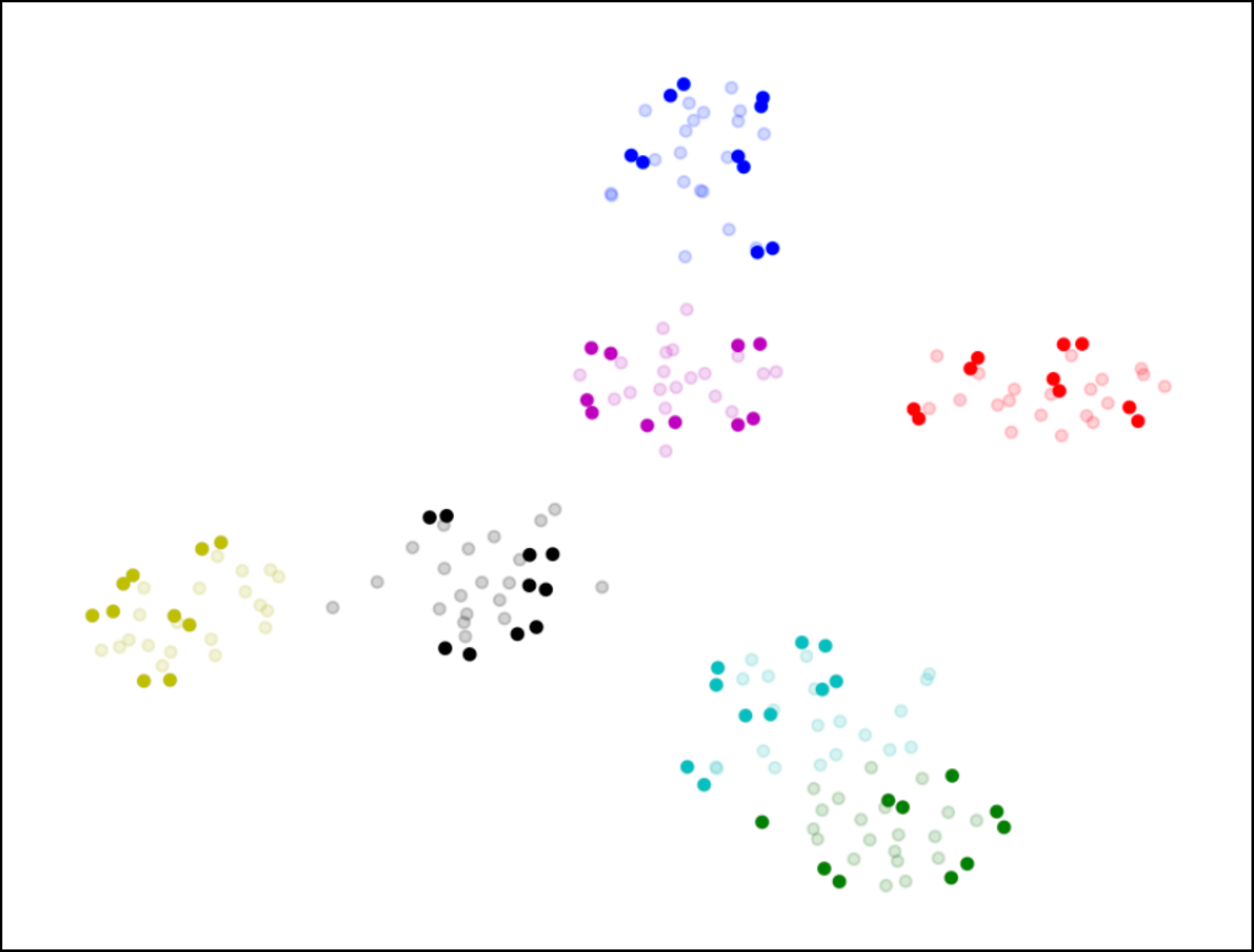}
\makebox[\Blank]{}
\includegraphics[width=\W,height=\H]{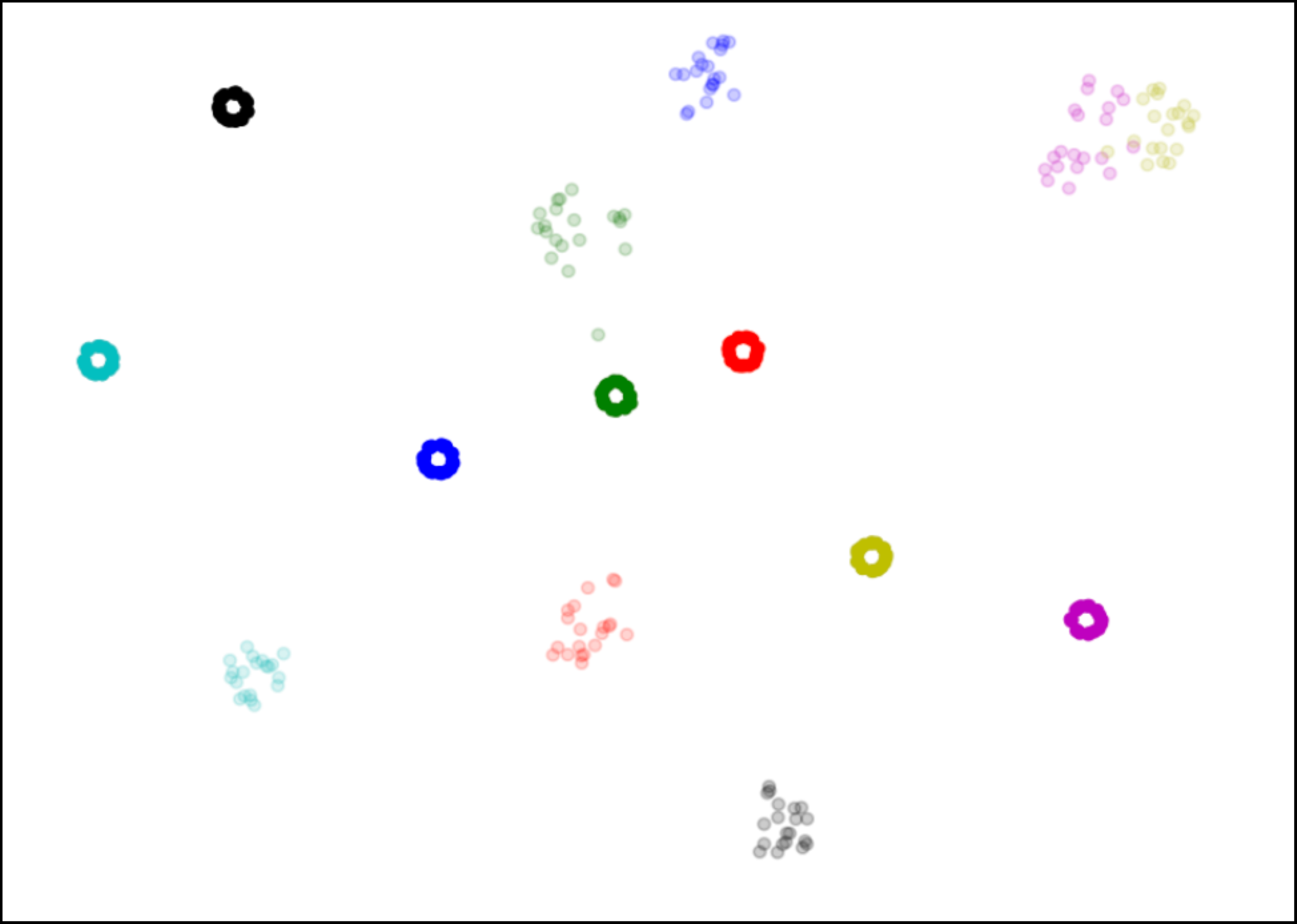}
   \makebox[\subboxsize]{(a) Ours - Base Classes}
   \makebox[\Blank]{}
    \makebox[\subboxsize]{(b) Ours - Novel Classes}
    \makebox[\Blank]{}
    \makebox[\subboxsize]{(c) $\Delta$-encoder - Novel Classes}
  \\
   \caption{Visualization of the original features (marked as dark points) and the augmented features (marked as transparent points) on the CUB dataset using t-SNE. (a) Base real features and features augmented by our method. (b) Novel real features and features augmented by our method. (c) Novel real features and features augmented by the $\Delta$-encoder.  
   Our augmented features better mimic the distribution of real features.
  }
  \label{fig:feats_tsne}
\end{figure*}

\subsection{Comparisons to other methods to model intra-class variance}

We assume that the intra-class variance can be modelled with an isotropic multivariate Gaussian distribution in a latent space.
%
%
In this section, we compare this method with other methods that model the intra-class variance including Gaussian mixture variational autoencoder (GMVAE) \cite{gmm_vae}, covariance matrix \cite{feature_transfer}, and a baseline model where we do not disentangle intra-class variance features from class-discriminative features.

Tab. \ref{tab:distribution_comparison} summarizes the results. The first row shows the results for GMVAE. This method enforces that the latent space is divided into distinct clusters for different classes. However, for this model, the accuracy drops by $6.15\%$ and $2.13\%$ for the 1-shot setting and $5.02\%$ and $0.50\%$ for the 5-shot setting on the CUB and NAB datasets respectively. The results align with our assumption that the intra-class variance is shared across different classes. Thus, enforcing a multi-modal prior distribution would lead to performance degradation. 

The second row shows the results for the method proposed in \cite{feature_transfer} based on covariance matrices.
%
%
%
%
%
%
Specifically, this method assumes a Gaussian prior on the distribution of intra-class variability across different classes which can be transferred from the base classes to the rare classes.
However, instead of modelling the distribution by variational inference, \cite{feature_transfer} uses a covariance matrix to estimate the feature variance distribution.
%
Here we apply this method on our extracted features to generate additional features on the CUB and NAB datasets under both 1-shot and 5-shot settings.
%
%
Compared with the non-parametric estimate of the Gaussian distribution, modelling intra-class variance via variational inference in an end-to-end manner brings $6.03\%$ and $1.91\%$ improvement for the  1-shot setting and $3.64\%$ and $1.03\%$ improvement for the 5-shot setting on the CUB and NAB dataset respectively.
Last, we provide the results for our method without  feature disentanglement, denoted as ``No disentanglement'' in the third row. In spirit, this model is similar to \cite{variational_fewshot} which models each point as a distribution via variational inference. Given a new sample, we augment it via sampling repeatedly from the estimated mean and variance. 
Without feature disentanglement and explicit modelling of the intra-class variance, this model does not achieve comparable results compared to other methods. 

 \begin{table}[!h] 
    \centering
  \resizebox{0.45\textwidth}{!}{%
    \begin{tabular}{l|cc|cc}
      \hline
      Intra-class distribution model & \multicolumn{2}{c|}{CUB} & \multicolumn{2}{c}{NAB}   \\
      & 1-shot & 5-shot & 1-shot & 5-shot \\
      \hline
      Gaussian Mixture Model \cite{gmm_vae}        & 75.16  & 86.46  & 86.49 & 94.72 \\
      Covariance Matrix \cite{feature_transfer}   & 75.28  & 87.84 & 84.71 & 94.19  \\
      No disentanglement \cite{variational_fewshot}           & 73.40  & 86.60  & 81.83  & 92.83  \\ 
      \hline
      Isotropic Gaussian (Proposed)            & \textbf{79.12}  & \textbf{91.48}  & \textbf{88.62}  & \textbf{95.22}  \\ 
      \hline
    \end{tabular}} \\ 
    \caption{Few-shot classification accuracy on the CUB ~\cite{cub} and NAB \cite{nab} dataset in 1-shot and 5-shot setting with different methods to model intra-class variance.
    }
    \label{tab:distribution_comparison}%
  \end{table}

\subsection{Data Distribution Analysis}

We  compare the data distributions between the real data and the generated data from our method in comparison to other state-of-the-art data generation methods \cite{feature_transfer,delta-encoder}.  Here we measure the average intra-class variance, the distances between classes (inter-class distances), and the data clusterability via the Davies–Bouldin index (DBI) \cite{dbindex}.  
Specifically, the DBI for a cluster $i$ is calculated by:
\begin{equation} \label{DBI}
      \begin{split}
          DBI_{i} =  \max_{i \neq j}  \frac{Intra_{(i)} + Intra_{(j)}}{Inter_{(i,j)}}
      \end{split}
\end{equation}
where $Intra_{(i)}$ is the intra-class variance of cluster $i$, calculated by taking the average of squared deviations from the class center.
$Inter_{(i,j)}$ is the distance between the two class centers of clusters $i$ and $j$.
The lower the value of the DBI, the better the separation between the clusters and the ``tightness'' inside the clusters.

Tab. \ref{tab:dbindex_comparison} shows the average values of the intra-class variance, inter-class distances,  and the DBI (denoted as $D_{intra}$, $D_{inter}$, and $DBI$ respectively) across all novel classes of the CUB dataset. The inter-class distances are averaged across all pairs of classes.
%
As shown in the table, features from the support set exhibit smaller intra-class variance compared to features from all data. All methods augment features from the support set. 
Interestingly, both  sets of generated features using the method proposed in \cite{feature_transfer} and the $\Delta$-encoder\cite{delta-encoder} decrease intra-class variance. On the other hand, the set of features augmented by our method closely approximate the data distribution of the set of all real features.


%
%
%
 \begin{table}[!h] 
    \centering
  \resizebox{0.4\textwidth}{!}{%
    \begin{tabular}{l|c|c|c}
      \hline
      & $D_{intra}$ & $D_{inter}$ & $DBI$ \\
      \hline
      Support data (5 samples)         & 21.52  & 32.77 & 2.21  \\
      All data    & 28.97 & 35.89 & 3.02  \\
      \hline
      Covariance matrix \cite{feature_transfer}  & 17.98  & 35.24 & 1.79  \\
      Encoder-based Model  \cite{delta-encoder} & 10.34  & 11.69  & 1.77    \\ 
      \hline
      Ours  & 27.27  & 34.12  & 2.53    \\ 
     \hline
    \end{tabular}} \\ 
    
    \caption{\textbf{Data Distribution analysis for different sets of features}. We augment features using our method and other data generation method based on covariance matrices \cite{feature_transfer} or the $\Delta$-encoder \cite{delta-encoder}. All methods augment features from the support set (first row). 
    }
    \label{tab:dbindex_comparison}%
  \end{table}


Fig. \ref{fig:feats_tsne} demonstrates how real samples and generated samples from our method are distributed in a 2D space in comparison with the $\Delta$-encoder \cite{delta-encoder} using t-SNE \cite{Maaten2008VisualizingDU}. The original features are marked as light colors while the augmented features are marked as dark colors. Different colors denote different classes. The visualization for the base classes with the augmented features from our method is shown in Fig. \ref{fig:feats_tsne}a. Fig. \ref{fig:feats_tsne}b visualizes the real features and the generated features of our method for the novel classes. Our method generates samples that follow closely the real samples. The visualization for the novel classes and the generated features from the $\Delta$-encoder is shown in Fig. \ref{fig:feats_tsne}c. As can be seen, the generated data from each novel class forms into a new cluster and does not lie close to the  actual data points. 
%

\section{Conclusion} 
   We have proposed a simple, yet effective, feature augmentation method via feature disentanglement and variational inference  to address the data scarcity problem in few-shot fine-grained classification.
     The generated features enlarge the intra-class variance for novel set images while preserving the class-discriminative features.
     The consistent performance improvement with the increase of the number of augmented samples suggests that the learned features are meaningful and nontrivial.
   The higher accuracy compared with other data augmentation based methods further demonstrate the superiority of our method.
     While this work mainly focuses on few-shot recognition problems, a promising future direction is to apply the feature transfer idea to other data-scarce or label-scarce tasks.
     
\newcommand{\myheading}[1]{\vspace{1ex}\noindent \textbf{#1}}
\myheading{Acknowledgements.} 
This work was partially supported by Zebra Technologies, the Partner University Fund, the SUNY2020 ITSC, and a gift from Adobe. 

{\small
\bibliographystyle{ieee_fullname}
\bibliography{egbib,neuron_science}
}

\end{document}